\documentclass[runningheads,a4paper]{llncs}

\usepackage{amssymb}
\usepackage{graphicx}

\usepackage{url}

\urldef{\mailsa}\path|{violos, ikom, papadop}@iti.gr|
\urldef{\mailsb}\path|{it218116}@hua.gr|
\newcommand{\keywords}[1]{\par\addvspace\baselineskip
\noindent\keywordname\enspace\ignorespaces#1}

\begin{document}

\mainmatter  


\title{Frugal Machine Learning for Energy-efficient, and Resource-aware Artificial Intelligence}

\titlerunning{Artificial Intelligence, Data and Robotics: Frugal Machine Learning}

%
%
\author{John Violos%
\and {Konstantina-Christina Diamanti \and Ioannis Kompatsiaris \and Symeon Papadopoulos}}
\authorrunning{Artificial Intelligence, Data and Robotics: Frugal Machine Learning}

\institute{Information Technologies Institute, Centre for Research and Technology Hellas\\
6th km Charilaou-Thermi, 57001 Thessaloniki, Greece\\
\mailsa\\
Dept. of Informatics \& Telematics, Harokopio University, Athens, 177 78 Greece\\
\mailsb\\
}

%
%

\toctitle{Artificial Intelligence, Data and Robotics: Frugal Machine Learning}
\tocauthor{Authors' Instructions}
\maketitle

\begin{abstract}
Frugal Machine Learning (FML) refers to the practice of developing Machine Learning (ML) models that are efficient, cost-effective, and resource-aware. Approach in this field aim to achieve satisfactory performance while minimizing the computational resources, time, energy and data required for training and inference. This chapter examines recent advancements, applications, and ongoing challenges in FML, highlighting its significance in smart environments that integrate edge computing and IoT devices, often constrained by limited bandwidth or high latency. Key technological enablers are explored, including model compression, energy-efficient hardware, and data-efficient learning techniques, highlighting their role in promoting sustainability and accessibility in Artificial Intelligence (AI). Furthermore, it provides a comprehensive taxonomy of frugal methods, discusses case studies across diverse domains, and identifies future research directions to drive innovation in this evolving field.

\end{abstract}

\section{Introduction}
\label{sec:Introduction}
FML aims to make AI more accessible and sustainable, particularly in resource-constrained environments such as smart environments, IoT systems, and embedded platforms \cite{bimpas_leveraging_2024}. Unlike traditional ML approaches that assume abundant resources, FML focuses on creating models that perform effectively on devices with limited computational power, data, memory, and energy \cite{kinnas_reducing_2025}. This is increasingly important as more applications move towards edge computing, which requires real-time processing on low-power devices \cite{llisterri_gimenez_-device_2022}.

FML also contributes to sustainability by reducing energy consumption, making AI development more environmentally friendly, while democratizing access to powerful models in settings with limited infrastructure \cite{li_evaluating_2016}. It enables scalable and cost-effective deployment across industries such as IoT and Edge computing \cite{violos_towards_2024}, wearable and mobile AI \cite{tan_deep_2023}, autonomous systems \cite{capra_hardware_2020} and cybersecurity \cite{yin_igrf-rfe_2023}, benefiting regions with constrained resources.

While closely related to Tiny Machine Learning (TinyML), FML and TinyML differ in focus. FML emphasizes reducing costs associated with data, computation, and time, while maintaining model performance across various ML tasks. TinyML, on the other hand, is specifically designed for ultra-low-power devices like micro-controllers and edge devices, optimizing models for minimal memory, processing power, and energy consumption. Though FML is broader and cost-driven, TinyML is specialized for small, resource-limited hardware platforms.

Key technological advancements have fueled the rise of FML, including the shift from centralized cloud computing to edge computing. This transition has driven the demand for lightweight models that can operate on devices with limited processing power. Techniques such as model pruning \cite{cheng_survey_2024}, quantization \cite{cherniuk_quantization_2024}, and knowledge distillation \cite{gou_knowledge_2021} have enabled significant reductions in model size with minimal performance loss, making them suitable for embedded systems.
Energy-efficient hardware, such as specialized AI accelerators like GPUs \cite{li_evaluating_2016} and TPUs \cite{pandey_greentpu_2019}, also play a crucial role in optimizing ML algorithms for low-power devices. These developments support frugal algorithms by reducing the energy consumption required for AI computations.
Additionally, techniques such as feature selection \cite{cancela_e2e-fs_2023} and data selection \cite{kinnas_selecting_2025} help retain the most informative data points, reducing the amount of training data required without compromising the accuracy of ML models.

\begin{table}[]
\begin{tabular}{|l|l|}
\hline
Objectives       & Description                                                                                                                                                                                                                                                           \\ \hline
Data Size        & Minimizing the amount of (labeled) data for the training processes.                                                                                                                                                                                                     \\ \hline
Computation      & Reducing the computational load by building lightweight models.                                                                                                                                                                                                    \\ \hline
Energy           & Minimizing energy consumption during training and inference.                                                                                                                                                                                         \\ \hline
Scalability      & \begin{tabular}[c]{@{}l@{}}Designing models that scale across various devices in constrained  \\ environments with different levels of resource availability.\end{tabular}                                                                                          \\ \hline
Monetary Cost    & \begin{tabular}[c]{@{}l@{}}Optimizing computational, memory, and data resources to reduce \\ the overall expense of deploying and maintaining ML systems.\end{tabular}                                                                                                  \\ \hline
Model size & Decreasing the number of parameters (weights and biases) in ANNs.                                                                                                                                                                                           \\ \hline
Compression & \begin{tabular}[c]{@{}l@{}}Reducing the time and resources required to compress, fine-tune,  \\ and validate a compressed ANN.\end{tabular}                                                                                                                            \\ \hline
Trade-offs      & \begin{tabular}[c]{@{}l@{}} FML balances the above objectives with performance, ensuring \\ efficient operation without compromising model accuracy.
\end{tabular} \\ \hline
\end{tabular}
\label{tab:Objectives}
\caption{Objectives of FML}
\end{table}

The chapter relates to the efficiency and sustainability of the AI, Data and Robotics Partnership \cite{curry_partnership_2022}.
Its main contribution is a comprehensive survey of FML, structured to ensure clarity and accessibility, helping readers easily understand and engage with the topic. 
Section \ref{Sec:Taxonomy} makes a taxonomy of most prominent FML methods, models and approaches. 
Section \ref{Sec:Usecases} presents various use cases and applications of FML. 
Section \ref{Sec:Observations} gives the observations about technologies and methodologies.
Section \ref{Sec:Challenges} casts a critical eye in FML discussing limitations, research gaps and providing suggestions for future work. 
Finally, we conclude the article in Section \ref{Sec:Conclusions}.

\section{A Taxonomy of FML Methods}
\label{Sec:Taxonomy}
Frugality in ML models can be achieved through several approaches. The most common approach involves compressing large ANNs into smaller (Subsection \ref{Sub:Compression}). Another approach leverages specialized hardware accelerators such as GPUs, TPUs, FPGAs, and NPUs to enhance computational efficiency while lowering energy consumption (Subsection \ref{Sub:HW}). Algorithm optimization focuses on selecting lightweight ML techniques or refined versions of existing models to improve efficiency (Subsection \ref{Sub:Algorithm}). Feature selection streamlines input data by eliminating non-essential features that have minimal impact on model performance (Subsection \ref{Sub:Feature}). Similarly, data sampling reduces training data by discarding entire samples that contribute little to the learning process (Subsection \ref{Sub:Sampling}). Furthermore, inverse modeling techniques facilitate the rapid determination of hyperparameter values that influence ML model performance (Subsection \ref{Sub:Hyperparameter}).

\begin{figure*}[ht!]
\centering
\includegraphics[width=1.0\textwidth]{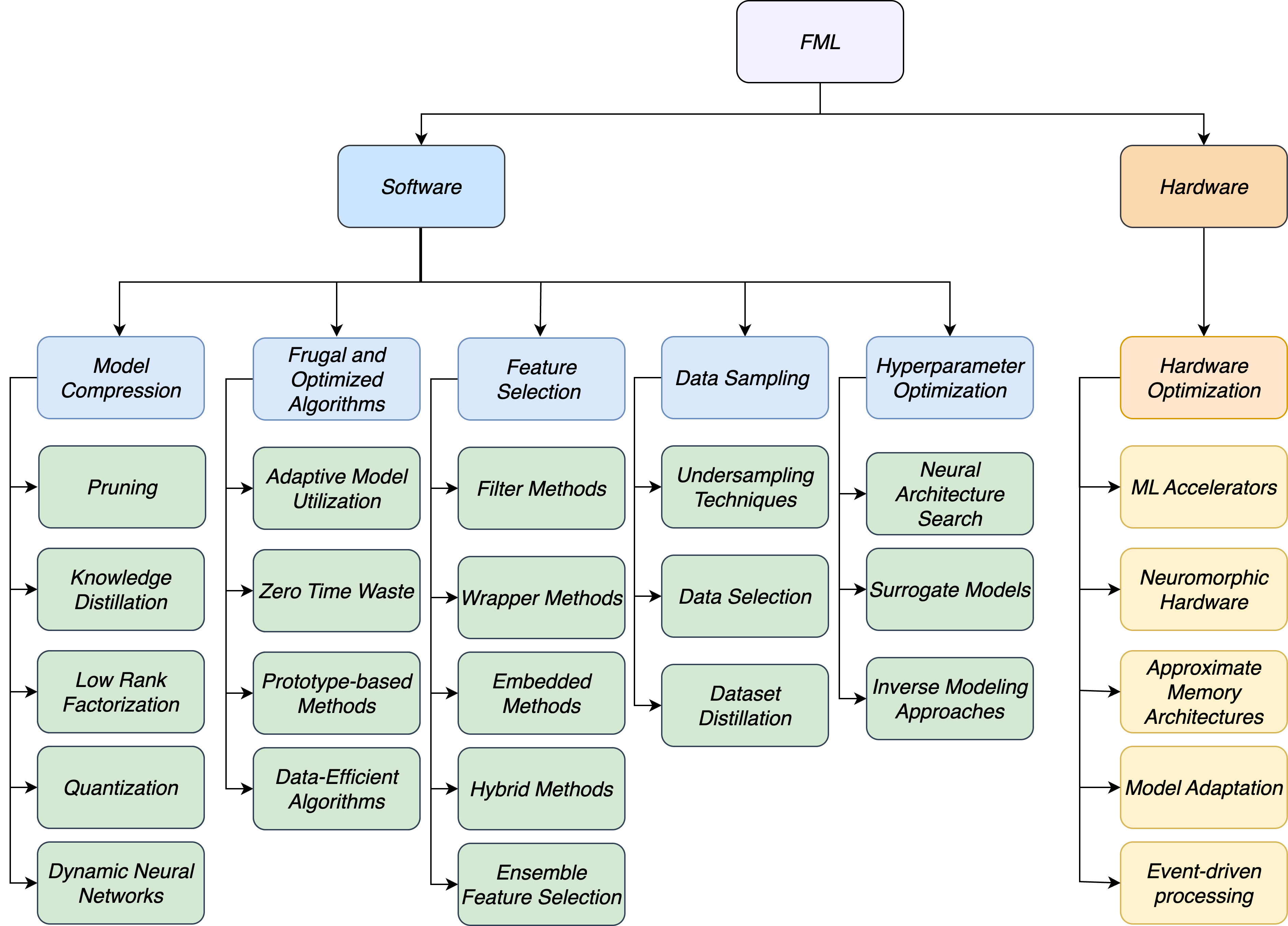}
\caption {Illustration of the FML Taxonomy}
\label{fig:technologies}
\end{figure*}

\subsection{Model Compression}
\label{Sub:Compression}
Model compression in FML refers to techniques that reduce the size and computational complexity of ANNs \cite{karathanasis_comparative_2025,}. This can be achieved through methods such as pruning, quantization, knowledge distillation, and low-rank factorization, or Dynamic Neural Networks that dynamically shrink the size of the ANN based on available computational resources.

\subsubsection{Pruning} Pruning is a widely used FML technique that reduces the size and complexity of ANNs by eliminating redundant or insignificant parameters while maintaining overall performance \cite{cheng_survey_2024}. This approach is particularly valuable for deploying deep learning models in environments with limited computational and memory resources, such as mobile devices and embedded systems. Pruning typically follows a two-step process: first, an overparameterized network is trained to achieve high accuracy; then, unnecessary weights, neurons, or filters are identified and pruned \cite{liebenwein_lost_2021}. To mitigate the potential accuracy loss from pruning, a fine-tuning phase is often incorporated to restore the model’s performance.

The two main categories of pruning is structured and unstructured. Structured pruning involves removing entire structures such as neurons, filters, channels, or layers, thereby maintaining a regular architecture and facilitating efficient hardware implementation \cite{he_structured_2024}. This approach contrasts with unstructured pruning, which eliminates individual weights, leading to irregular sparsity patterns that can be challenging for standard hardware to handle \cite{zou_convolutional_2018}. An example of structured pruning is filter pruning in CNNs, where entire filters are removed based on specific criteria. For instance  removing filters with small L1-norm values to compress CNNs \cite{li_pruning_2017}. A third approach is dynamic pruning which refers to methods that adjust the network's structure during training or inference, allowing the model to adaptively prune and regrow connections based on certain criteria \cite{chen_dygnn_2021}. 

An important pruning technique is the magnitude-based pruning, which assesses the importance of parameters based on their magnitude, with smaller weights typically considered less critical \cite{li_stage-wise_2024}. Sensitivity-based pruning goes further by analyzing the effect of removing specific weights or neurons on the model’s loss function, often using techniques like Taylor series approximations for precise pruning decisions \cite{molchanov_pruning_2017}. Additionally, clustering and similarity-based methods identify and group highly similar or duplicate parameters to reduce redundancy while retaining the network’s representational power \cite{song_filter_2023}. 

Recent advancements in pruning methodologies include optimization-driven approaches such as filter pruning and dynamic network surgery \cite{guo_dynamic_2016}. These methods adaptively prune filters or feature maps during training to ensure sustained performance. The introduction of concepts like the Lottery Ticket Hypothesis has further improved pruning strategies by uncovering high-performing sub-networks within large trained models \cite{frankle_lottery_2019}. Identified through iterative pruning and fine-tuning, these sub-networks have achieved remarkable reductions in model size while maintaining accuracy in tasks such as image recognition and natural language processing. 

\subsubsection{Knowledge Distillation (KD).} Knowledge Distillation is a model compression technique designed to transfer knowledge from a large teacher model to a smaller, lightweight student model \cite{gou_knowledge_2021}. This approach enables the student model to mimic the behavior of the teacher model while achieving a comparable level of performance. KD is especially advantageous in scenarios with constrained computational resources, such as IoT and edge computing devices \cite{violos_towards_2024}. The process involves the teacher model generating ``soft targets'' which are class probabilities that encapsulate relational information among different classes. These soft targets guide the student model during training, allowing it to achieve high performance with significantly fewer parameters and reduced computational complexity. Recent advancements, such as self-distillation and multi-teacher frameworks, have further extended KD’s utility, enabling efficient optimization and scalability for diverse tasks \cite{pham_collaborative_2023}.

The KD process comprises three key components: knowledge, the distillation algorithm, and the teacher-student architecture \cite{tsanakas_light-weight_2024}. Knowledge can be derived from multiple facets of the teacher model, including the final output layer (response-based knowledge), intermediate feature representations (feature-based knowledge), or the relationships among data points (relation-based knowledge) \cite{yang_categories_2023}. The distillation algorithm typically uses loss functions, such as Kullback-Leibler divergence, to align the outputs of the student model with those of the teacher model. A teacher-student architecture should maintain a balanced capacity between the two models to facilitate effective knowledge transfer \cite{gou_knowledge_2021}.

\subsubsection{Low-Rank Factorization}
Low-rank factorization is a technique used in ANN compression, where large weight matrices are approximated by smaller, low-rank matrices \cite{cai_learning_2023}. This approach leverages the fact that many weight matrices in neural networks contain redundancies and can be efficiently represented by decomposing them into lower-dimensional forms. One common method of low-rank factorization is Singular Value Decomposition (SVD), which decomposes a matrix into three smaller matrices, and the rank is reduced by retaining only the largest singular values \cite{yang_learning_2020}. 
Another method is Matrix Decomposition, which includes techniques like CUR decomposition and Alternating Least Squares (ALS) \cite{park_curing_2025}. These techniques factorize the original matrix into smaller components, further reducing its dimensionality and improving computational efficiency. 

\subsubsection{Quantization} Quantization reduces model size and computational requirements by lowering the number of bits needed to represent each parameter. This process can reduce the precision of model weights and activations from 32-bit floating-point to 8-bit integers \cite{nalepa_towards_2020}. Although post-training quantization may lead to performance loss due to reduced precision, quantization-aware training (QAT) mitigates this by integrating quantization effects during training, allowing models to adjust to the process and maintain higher accuracy \cite{menghani_efficient_2023}. 

These compression techniques, though varied in their approaches, can be combined to maximize efficiency. For instance, pruning and quantization can be applied together to simultaneously reduce model size and inference time \cite{tabani_improving_2021}. 
Quantized distillation leverages distillation during the training process, by incorporating distillation loss, expressed with respect to the teacher, into the training of a student network whose weights are quantized to a limited set of levels \cite{polino_model_2018}. Quantization also has been combined with low-rank factorization by performing Canonical Polyadic  decomposition with factor matrices constrained to a lower precision format (e.g., INT8/INT6/INT4) \cite{cherniuk_quantization_2024}.

\subsubsection{Dynamic Neural Networks} 
Typically, ANNs are designed as static models with fixed architectures and computation patterns. Furthermore, Dynamic Neural Networks (DyNNs) have been developed to introduce flexibility by allowing adjustable parameters such as depth, width, and activation pathways, which can change dynamically during inference or training \cite{liu_incremental_2023}. These networks possess several key features that enhance their efficiency and adaptability. They utilize adaptive computation to adjust the number of layers, neurons, or operations based on input complexity. Conditional execution enables selective activation of network components, such as skipping layers or neurons, based on learned policies \cite{wang_skipnet_2018}. Additionally, input-dependent processing ensures efficient resource allocation by dedicating fewer computations to simpler inputs while allocating more resources for complex ones \cite{liu_dynamic_2018}. As a result, DyNNs offer energy efficiency, reducing unnecessary computations and optimizing power consumption—making them particularly beneficial for real-time applications and deployment on resource-constrained edge devices.

DyNNs can be categorized into different types based on how they adapt their structure and computation. Spatially adaptive networks modify their architecture at different spatial locations in an image, such as using dynamic convolution to process regions with varying complexity more efficiently \cite{verelst_dynamic_2020}. Temporally adaptive networks adjust the number of timesteps in recurrent architectures like RNNs or transformers, allowing them to allocate computational resources based on the temporal dependencies of the input sequence \cite{varghese_transformerg2g_2024}. Routing-based networks leverage mechanisms like gating or reinforcement learning to selectively activate sub-networks, as seen in Mixture of Experts models, which dynamically route inputs to the most relevant network components \cite{zhou_mixture--experts_2022}. Lastly, self-pruning networks enhance efficiency by learning to remove redundant weights or neurons during training or inference, reducing computational cost while maintaining performance \cite{elkerdawy_fire_2022}. These dynamic architectures enable efficient and flexible deep learning models suited for a variety of tasks.

One specific category of DyNNs are the Slimmable Neural Networks (SNNs).
SNNs are designed to dynamically adjust their architecture based on available computational resources, offering flexible configurations in terms of depth (number of layers) and width (number of neurons per layer) \cite{yu_slimmable_2018}. During inference, SNNs can select different model configurations depending on input complexity and hardware constraints, making them adaptable for a variety of environments \cite{li_dynamic_2021}. This scalable computation allows the network to use lightweight architectures for simpler tasks and scale up to deeper or wider configurations for more complex tasks. SNNs are trained to support all possible configurations, with mechanisms that enable parts of the network to be turned on or off as needed.

In addition to depth and width slimming, channel slimming reduces the number of channels (filters) in each convolutional layer of CNNs  \cite{qiu_slimconv_2021}. Similarly, block slimming applied in modular networks enables the dynamic selection of blocks, such as residual or attention blocks, based on the task requirements or available resources \cite{li_dynamic_2021}. Last, Skip Connections and Residual Slimming adjusts the connections between layers, skipping unnecessary ones to further reduce computation \cite{xu_development_2024}. Each type of slimming contributes to enhancing model flexibility and efficiency, depending on the computational demands and input complexity. Similar to SNNs, Adaptive Neural Networks adjust computational complexity based on resource constraints, but while SNNs switch between predefined network widths, Adaptive Neural Networks dynamically modify their structure per input using mechanisms like early exiting or conditional activation \cite{lee_low-power_2013}.

\subsection{Frugal and Optimized Algorithms}
\label{Sub:Algorithm}
This category encompasses methods that dynamically adjust their computational demands based on task complexity. It includes approaches that minimize computational waste in early-exit strategies, methods inherently designed to be lightweight, such as prototype-based techniques, and those specifically developed to operate effectively with limited data.

\subsubsection{Adaptive Model Utilization (AMU)} 
AMU focuses on using computationally intensive DNNs only when necessary while relying on lightweight DNNs or methods otherwise. In computer vision, one proposed approach activates DNNs selectively to detect new objects or re-identify those that undergo significant appearance changes \cite{apicharttrisorn_frugal_2019}. During intervals between DNN executions, lightweight methods track detected objects efficiently without compromising performance. A more efficient AMU approach employs a dual-network architecture that dynamically alternates between a lightweight CNN for simpler tasks and a larger CNN for more complex tasks, depending on whether the confidence score of the smaller CNN drops below a predefined threshold \cite{park_biglittle_2015}. Furthermore, instead of using a lightweight and a larger CNN it has been proposed an AMU method that works on the synergy of two complementary lightweight CNNs and a memory component \cite{kinnas_reducing_2025}.
In this method each network compensates for the other's weaknesses in predictions. If the first CNN has low confidence in its output, the second CNN is activated to improve accuracy. This method also using a memory component that stores past classifications, avoids redundant computations.

\subsubsection{Zero Time Waste (ZTW)} 
Zero Time Waste (ZTW) is an optimization method for early exit neural networks that minimizes computational waste by reusing information from previous internal classifiers (ICs) rather than discarding their predictions \cite{wojcik_zero_2023}. Traditional early exit methods allow certain ICs to return predictions early for easy-to-classify samples, reducing overall inference time \cite{rahmath_p_early-exit_2024}. However, when an IC does not meet the confidence threshold to exit, its computations are wasted. ZTW addresses this inefficiency by introducing cascade connections, which pass the outputs of earlier ICs to later ones, allowing each classifier to refine previous predictions rather than starting from scratch. 

\subsubsection{Prototype-based Methods} Prototype-based methods have been characterized as frugal because they rely on a small set of representative prototypes rather than memorizing or optimizing over large datasets. They work by comparing data points with the representative examples of each class called prototypes \cite{cholet_framework_2023}. 
Incremental Neural Networks are prototype-based neural classifiers that dynamically adjust and expand their structure by incorporating new prototypes as data is progressively introduced \cite{kawewong_fast_2011}. In this way they enable efficient learning without the need for full retraining, even when integrating data from different data generative distributions. 
Adaptive Resonance Theory models also belong to the prototype-based methods by comparing the input data to existing patterns \cite{grossberg_nonlinear_1988}. If the new input is similar to a stored pattern then the existing pattern is updated. Otherwise, a new category is created.

\subsubsection{Data-Efficient Algorithms}  
Data-efficient algorithms are designed to optimize learning processes by reducing the need for large amounts of labeled data \cite{adadi_survey_2021}. These algorithms employ various strategies to enhance learning efficiency, making them particularly suitable for data-scarce environments. Semi-supervised learning is a key method, combining labeled and unlabeled data to improve learning without extensive labeling \cite{sarantos_enabling_2025}. Techniques such as self-training, co-training, and graph-based methods are employed to iteratively label and train on unlabeled data \cite{tu_frugal_2021}. Additionally, generative models and transformation-equivalent representations are utilized to learn robust features, further reducing the need for large datasets.

\subsection{Feature Selection}
\label{Sub:Feature}
Feature selection reduces data dimensionality by identifying the most relevant features. Techniques such as filter, wrapper, embedded, hybrid, and ensemble methods decrease computational and memory demands. By eliminating redundant or irrelevant features, they enhance input efficiency in traditional ML models.

\subsubsection{Filter Methods} Filter methods select relevant features independently of any ML model by using statistical measures to assess their importance. These methods operate as a pre-processing step, ranking features based on their correlation with the target variable. For example, the Symmetrical Uncertainty ranks features based the association between each feature and the target variable using entropy calculations \cite{senthamarai_kannan_novel_2010}. RELIEF, another filter method, evaluates feature importance by examining the values of the nearest instances of the same and different classes, capturing feature dependencies effectively \cite{urbanowicz_relief-based_2018}. 

\subsubsection{Wrapper Methods} Wrapper methods evaluate feature subsets by training and testing a specific ML  model on different combinations of features. These methods use model performance as the criterion for selecting the best subset of features. 
A widely-used wrapper method is Forward Selection, which begins with an empty feature set and progressively adds the most significant feature at each step until performance reaches its peak \cite{saha_correlation_2020}. In contrast, Backward Elimination starts with all available features and removes the least important ones at each step, guided by the model's performance \cite{nguyen_filter_2014}.

\subsubsection{Embedded Methods} 
Unlike filter methods, which evaluate features independently of the model, and wrapper methods, which rely on separate evaluation processes, embedded methods integrate feature selection directly into model training. These methods have been applied to traditional ML models such as SVMs and ANNs. In SMVs, feature selection has been formulated as a min-max optimization problem that tries to balance model complexity and classification accuracy \cite{jimenez-cordero_novel_2021}. By leveraging duality theory, this method reformulates the problem into a solvable non-linear optimization framework. In ANNs, an embedded method has been introduced, incorporating a binary mask layer that dynamically selects the most relevant features during training using gradient-based optimization \cite{cancela_e2e-fs_2023}. By applying regularization constraints to enforce sparsity, the embedded method forces the selection of a predefined maximum number of features, ensuring computational efficiency while maintaining accuracy.

\subsubsection{Hybrid Methods} Hybrid methods combine the strengths of filter and wrapper methods to balance computational efficiency and model performance. 
The hybrid approach combines the speed of filter methods with the relevance search capabilities of wrapper methods \cite{yin_igrf-rfe_2023}. In the first phase, two filter methods, information gain and random forest, are utilized to reduce the feature subset search space efficiently. In the second phase, a wrapper method applies recursive feature elimination, further refining the feature set while preserving the most relevant features and considering feature similarities. 

\subsubsection{Ensemble Methods}
Ensemble Feature Selection entails combining multiple feature selection methods and aggregating their results, often using weighted voting \cite{seijo-pardo_ensemble_2017}.  Homogeneous ensemble methods apply the same selection algorithm to different training data subsets, whereas heterogeneous ensemble methods integrate various selection algorithms on the same dataset. This approach enhances stability by reducing the risk of selecting inconsistent feature subsets and ensures a more reliable feature ranking.

\subsection{Data Sampling }
\label{Sub:Sampling}
Data sampling, initially used to address imbalanced datasets, is now gaining attention in FML for its ability to optimize resource efficiency. Techniques such as under-sampling significantly reduce the size of training data, resulting in shorter training times, lower computational costs, and decreased resource consumption. By selecting the most informative data samples, data sampling aims to preserve algorithm performance, ensuring that models remain effective even with a more compact dataset.

\subsubsection{Under-sampling Techniques} 
Random under-sampling in the majority class is a straightforward approach but may lead to the loss of valuable information. Edited Nearest Neighbors, eliminates majority class samples that are misclassified by their nearest neighbors, thereby enhancing the classifier's ability to distinguish between classes \cite{mohammed_machine_2020}.
A different approach identifies and removes samples in the majority class that are close to minority class samples \cite{wernerdevargas_imbalanced_2023}.   

\subsubsection{Data Selection} 
Data selection methods use ranking criteria to assess and prioritize samples for training an ANN or for knowledge distillation. Manifold Learning-based data selection works by projecting high-dimensional volumetric  data into a lower-dimensional plane, where each point represents a volumetric data instance. Then, clustering techniques group the data points into distinct classes, from which the training dataset is sampled \cite{chen_manifold_2018}. 
An entropy-based data selection method has also been proposed that identifies smaller subsets from the original dataset by focusing on instances that retain the most informational value \cite{kinnas_selecting_2025}. This method calculates entropy using the logit vectors of data samples and selects those with the highest entropy.

\subsubsection{Dataset Distillation} 
Dataset distillation compresses knowledge from a large dataset into a much smaller synthetic dataset by optimizing it so that models trained on it perform comparably to those trained on the original data \cite{wang_dataset_2020}.
The process involves initializing a small synthetic dataset, training a neural network on it, and optimizing the data using the following three objectives \cite{yu_dataset_2023}: a)  performance matching, which ensures models trained on synthetic data achieve similar accuracy as those trained on real data; b) parameter matching, which matches the internal parameters of networks trained on real and synthetic data; and c) distribution matching, which ensures the synthetic data maintains statistical properties similar to the original dataset. The synthetic dataset is then refined through iterative steps to maximize its informativeness.

\subsection{Hyperparameter Optimization} 
\label{Sub:Hyperparameter}
ANNs and traditional ML models involve numerous hyperparameters related to both their architecture and training process. Identifying the optimal hyperparameters is often done through trial and error, requiring practitioners to test multiple combinations to achieve the most accurate model. Hyperparameter optimization aims to streamline this process by efficiently exploring the hyperparameter space, reducing the number of experiments needed to achieve the FML objectives. 

\subsubsection{Neural Architecture Search} 
Neural Architecture Search  is an automated technique for designing optimal neural network architectures that consists of three key components: a) search space, which defines the range of possible architectures, including factors like the number of layers, filter sizes, and connections \cite{salmani_pour_avval_systematic_2025}; b) search strategy, which determines how different architectures are explored, using methods such as reinforcement learning, genetic algorithms, or gradient-based optimization \cite{tsanakas_innovative_2021}; and c) evaluation method, which measures the performance of candidate architectures based on FML criteria like accuracy, efficiency, or computational cost.  

\subsubsection{Surrogate Models} 
Surrogate models are simplified, computationally efficient approximations of complex and expensive-to-evaluate functions, commonly used in ML to replace costly simulations \cite{cho_basic_2020}. The process begins with data collection, where a set of sample points is gathered from the original function or system. Using these samples, a surrogate model is constructed to approximate the function, providing a more efficient way to evaluate potential solutions \cite{diaw_efficient_2024}. This model is then used for optimization and exploration, identifying promising solutions while minimizing computational costs. To enhance accuracy, new samples from the original function can be iteratively incorporated, refining the surrogate model over time.

\subsubsection{Inverse Modeling Approaches.} 
Inverse modeling approaches are computational methods used to estimate unknown system parameters like ANNs architectures by analyzing observed outputs \cite{chan_inverse_2023}. These techniques work backward from the results, making them valuable for optimization and inference. The process begins by defining a forward model that captures the relationship between inputs and outputs. Observational data, such as real-world measurements, serve as the known outputs, and optimization or statistical inference techniques are then applied to estimate the unknown inputs by minimizing the discrepancy between the model’s predictions and actual observations. The estimated parameters are validated using independent data, and the model is refined iteratively to improve accuracy.

\subsection{Hardware Optimization}
\label{Sub:HW}
Hardware optimization involves the use of processing units designed for ML tasks, neuromorphic hardware, and efficient memory architectures. It also includes techniques that adapt ANNs to the deployed devices and event-driven processing methods, where ANN models operate only when necessary.

\subsubsection{ML Accelerators}
ML accelerators are specialized processing units engineered to support the objectives of FML.  Graphics Processing Units (GPUs), initially developed for graphics rendering, have become essential for ML due to their parallel processing capabilities, significantly minimizing energy consumption \cite{li_evaluating_2016} while accelerating neural network training and inference \cite{parnell_large-scale_2017}.
Additionally, Tensor Processing Units (TPUs) are specifically optimized for deep learning workloads, enabling low-computational matrix multiplications while consuming less power than GPUs and CPUs \cite{pandey_greentpu_2019}. 
Field-Programmable Gate Arrays (FPGAs) offer customizable ML acceleration, enabling optimized energy-efficient computations for embedded AI systems \cite{yan_survey_2024}. 
Neural Processing Units (NPUs), such as the Apple Neural Engine and Qualcomm Hexagon NPU, are dedicated AI accelerators designed for efficient deep learning inference on edge devices and mobile platforms \cite{tan_deep_2023}. 

\subsubsection{Neuromorphic Hardware}
Neuromorphic hardware consists of computing systems engineered to replicate the structure and function of the human brain by leveraging Spiking Neural Networks (SNNs) and specialized circuits that function similarly to biological neurons and synapses \cite{balaji_mapping_2020}. Unlike conventional von Neumann architectures, which distinctly separate memory and processing, 
neuromorphic computing systems both store and process data in individual neurons, resulting in lower latency and data transfer \cite{schuman_survey_2017}.
These systems process information using discrete spikes rather than continuous signals, reducing unnecessary computations. Additionally, asynchronous processing ensures computations occur only when necessary, minimizing energy consumption. By incorporating in-memory computing, neuromorphic hardware eliminates data transfer bottlenecks, improving speed and efficiency. 

\subsubsection{Approximate Memory Architectures}. 
Approximate Memory Architectures are memory systems designed to improve energy efficiency and performance by relaxing accuracy constraints in data storage and retrieval \cite{nguyen_approximate_2020}. An approximate memory architecture uses soft approximation through row-level refresh control and hard approximation via data truncation, adjusting DRAM refresh rates and data precision based on significance. 

\subsubsection{Model Adaptation Through Progressive Shrinking and Once-for-All Networks}
Progressive Shrinking is a training strategy that optimizes large neural networks by gradually training smaller sub-networks, reducing interference between different architectures while preserving accuracy. This approach enables efficient adaptation to diverse hardware constraints, allowing models to scale seamlessly across different devices \cite{capra_hardware_2020}. Once-for-All (OFA) Networks extend this concept by training a single, flexible model that can be specialized for different hardware setups without retraining. By decoupling training from neural architecture search and leveraging progressive shrinking, OFA Networks enable highly efficient deep learning deployment, particularly for resource-constrained environments like mobile and edge devices \cite{cai_once-for-all_2020}.

\subsubsection{Event-driven processing}
Event-driven processing is a computing approach where operations are executed only when specific events occur such as sensor inputs, user interactions, or data changes, rather than continuously processing data in a clock-driven manner \cite{kuhrt_efficient_2025}. This approach is particularly useful in hardware optimization because it reduces unnecessary computations, leading to lower energy consumption and improved efficiency.

\section{Use cases and Applications of Frugal Machine Learning}
\label{Sec:Usecases}
FML is particularly useful in scenarios requiring efficient AI with minimal computational resources, energy consumption, and data availability. Some key applications and use cases are briefly described in the following subsections.

\subsection{IoT and Edge AI}
FML enables lightweight AI models to run efficiently on low-power sensors, smart home devices, and  IoT systems, reducing reliance on cloud computing \cite{kinnas_reducing_2025}. By enabling on-device inference, these models improve real-time decision-making with local data, enhancing both privacy and control over the users'  data \cite{violos_towards_2024}. Furthermore, FML facilitates decentralized intelligence, allowing IoT and Edge AI systems to operate collaboratively through federated learning and peer-to-peer communication, enabling more robust, adaptive, and resilient AI applications in dynamic environments \cite{llisterri_gimenez_-device_2022}.

\subsection{Wearable and Mobile AI}
FML powers fitness trackers, activity recognition, and smartwatches, optimizing energy consumption for prolonged battery life \cite{tan_deep_2023}. By using compressed ANNs and efficient inference strategies, wearable AI can track heart rate, sleep patterns, and physical activity without frequent cloud interactions \cite{saha_correlation_2020}. This enables real-time health alerts and personalized recommendations while ensuring that devices last longer on a single charge.
Advanced techniques of FML and on-device learning further enhance the capabilities of wearables, allowing them to adapt and improve over time with minimal data exchange \cite{yu_slimmable_2018}.

\subsection{Autonomous Systems and Robotics}
In drones, self-driving cars, and warehouse robots, FML enhances efficiency by enabling real-time, low-power inference for navigation, obstacle detection, and decision-making \cite{kawewong_fast_2011}. 
Autonomous agents demonstrate robust adaptability across diverse environments, leveraging frugal learning to efficiently process and learn from sensor data.
Applications include autonomous delivery robots, smart transportation systems, and industrial automation, where energy-efficient AI ensures prolonged operation, better scalability, and reduced computational overhead while maintaining accuracy in complex environments \cite{capra_hardware_2020}.

\subsection{Healthcare and Medical AI}
FML supports low-resource diagnostics, remote health monitoring, and medical imaging in areas with limited infrastructure \cite{saha_correlation_2020}. By deploying lightweight AI models on portable medical devices and mobile health apps, doctors and caregivers can perform real-time patient monitoring, early disease detection, and predictive analytics even in remote locations. For example, compact AI models can analyze X-rays, detect anomalies in electrocardiogram readings, or provide AI-driven telemedicine support, improving accessibility to healthcare without requiring expensive computing resources \cite{lee_low-power_2013}.

\subsection{Ambient Intelligence}
FML enhances ambient intelligence in smart homes, smart cities, smart buildings, and smart agriculture, by enabling low-power AI models to process data locally on edge devices \cite{bimpas_leveraging_2024}. In smart homes and buildings, energy-efficient AI optimizes  heating, ventilation and air-conditioning, security, and appliance automation. Smart cities benefit from real-time traffic management, waste monitoring, and public safety enhancements. In smart agriculture, FML supports precision farming, pest detection, and irrigation control using low-power remote sensors.

\subsection{Bandwidth Constrained Systems}
FML facilitates AI applications in remote and rural areas by reducing data transmission needs and preserving bandwidth \cite{yu_dataset_2023}. By enabling on-device data processing and compression, AI models can extract key insights before sending only essential information to centralized servers. This is particularly useful in satellite-based IoT, remote health diagnostics, and off-grid communication networks, where connectivity is expensive or intermittent \cite{nalepa_towards_2020}. FML ensures efficient data utilization, making AI-powered services accessible even in bandwidth-constrained environments.

\subsection{Embedded AI for Consumer Electronics}
FML powers smart cameras, voice assistants, and AR/VR devices with optimized processing for real-time responsiveness \cite{apicharttrisorn_frugal_2019}. By enabling low-latency AI inference directly on devices, users experience faster interactions with technologies like speech recognition, facial authentication, and augmented reality overlays \cite{capra_hardware_2020}. Techniques such as quantized neural networks and hardware-aware optimizations allow complex AI tasks to be performed efficiently on compact, battery-operated devices, enhancing usability while extending battery life.

\subsection{Cybersecurity}
FML provides efficient threat detection in networked systems without excessive computational overhead \cite{yin_igrf-rfe_2023}. Lightweight AI models deployed in firewalls, intrusion detection systems, and endpoint security solutions can analyze network traffic, identify anomalies, and prevent cyber-attacks in real-time without straining system resources \cite{sarantos_enabling_2025}. Applications include fraud detection in financial transactions, malware classification on mobile devices, and anomaly detection in industrial networks, ensuring strong security while data requirements and computational load.

\section{Observations}
\label{Sec:Observations}
To achieve frugality, three main approaches exist: Input frugality, learning process frugality, and model frugality \cite{kinnas_selecting_2025}. Input frugality emphasizes minimizing the cost associated with data acquisition and feature utilization by using fewer training data or fewer features, driven by data availability, resource, or privacy constraints. Learning process frugality emphasizes minimizing the computational and memory resources required for training a model, often resulting in a less accurate but more resource-efficient model, driven by constraints such as limited computational power and training time. Model frugality emphasizes minimizing the memory, computational resources, and energy required to store and use a ML model, often at the expense of optimal prediction quality.

A key requirement for many ML models is the ability to adapt continuously to new tasks. Instead of retraining an entire ANN from scratch, a process that requires substantial computational power and energy, models can be updated incrementally, preserving learned representations in a more resource-efficient manner. Three core strategies support this approach: parameter regularization, knowledge distillation, and dynamic architecture design \cite{liu_incremental_2023}. Parameter regularization limits changes to critical weights, ensuring previously acquired knowledge is retained. Knowledge distillation allows newer models to inherit insights from earlier versions without storing entire datasets. Meanwhile, dynamic architectures enable networks to expand selectively, efficiently incorporating new tasks.

Lastly, FML often aligns with efforts to reduce the environmental impact of AI, sometimes called ``green computing''. Cutting back on resource use, including energy consumption, can help lower the overall carbon footprint of ML. However, standard methods to measure these gains and compare them across different projects are still limited. As research in FML continues, the field is expected to bring more practical, cost-friendly, and environmentally responsible ways to deploy AI in real-world applications.

\section{Open Challenges of Frugal Machine Learning}
\label{Sec:Challenges}
Despite significant progress in FML, several challenges must be overcome for broader adoption by ML researchers and practitioners. A key issue stems from the resource-constrained environments where FML is typically deployed, such as IoT and edge computing. Smart environments often handle noisy, redundant, or rapidly changing data, making efficient processing difficult. While techniques like under-sampling, feature selection and data selection help reduce input size and select the most appropriate data samples, they fail to fully compensate for limited processing resources. Additionally, the risk of catastrophic forgetting further complicates the task of ensuring stable and continuous learning, making long-term deployment even more challenging.

Another key challenge lies in balancing model interpretability and fairness with aggressive optimization. Compression techniques, while improving efficiency, often make it harder to understand and explain how a model arrives at its decisions. Additionally, these optimizations can unintentionally introduce bias, particularly when data distributions shift or certain subgroups become underrepresented. Maintaining transparency and fairness is crucial, especially in high-stakes applications like healthcare and finance. Striking the right balance between efficiency and trustworthiness remains an ongoing challenge in the adoption of FML.

One more open concern is the lack of standardized benchmarks and evaluation metrics capable of capturing various dimensions of resource constraints, from memory and inference latency to overall energy consumption and carbon footprint. Existing studies often focus on either accuracy or speed, employing different datasets and methodologies, which makes fair comparisons and reproducibility difficult. Establishing robust benchmarks that reflect real-world conditions, such as limited connectivity on edge devices or dynamic data streams, would facilitate direct comparisons of new techniques and accelerate progress. 

Finally, there is need for better guidance on when and how to deploy various FML strategies across different lifecycle stages, from hyperparameter searches to model deployment and maintenance. The sequence in which techniques like pruning and quantization are applied can significantly impact final performance, yet there are no standardized best practices or unified pipelines to streamline these processes. Future advancements could benefit from end-to-end tool chains that seamlessly integrate hyperparameter optimization, model compression, and hardware-aware deployment into a cohesive framework. Additionally, incorporating lifecycle cost assessments will be essential to ensure that reductions in model size or training time lead to genuine sustainability gains rather than superficial optimizations.

\section{Conclusions}
\label{Sec:Conclusions}
FML offers a resource-conscious approach to AI development by employing smaller models, efficient data use, and optimized algorithms. This enables strong performance even under constraints like limited processing power, scarce data, or low energy availability. 
Techniques such as feature selection, model compression, data sampling, and hardware optimization demonstrate FML’s adaptability across diverse settings, from mobile devices and IoT sensors to large-scale cloud systems. By promoting efficient resource use, FML supports AI deployment in low-bandwidth environments while advancing sustainability. As the field evolves, it is poised to drive innovation in accessible, efficient, and responsible ML solutions.

\section*{Acknowledgments}
  This work was funded by the European Union’s Horizon Europe research and innovation program under grant agreement No. 101121447 Covert and Advanced multi-modal Sensor Systems for tArget acquisition and reconnAissance (CASSATA) and innovation program under grant agreement No. 101120237 European Lighthouse of AI for Sustainability (ELIAS)

 \bibliographystyle{elsarticle-num}  
 \bibliography{References}

\end{document}